\documentclass[sigconf, nonacm]{acmart}

\AtBeginDocument{%
  \providecommand\BibTeX{{%
    \normalfont B\kern-0.5em{\scshape i\kern-0.25em b}\kern-0.8em\TeX}}}

\usepackage{algorithmic}
\usepackage[ruled,linesnumbered]{algorithm2e}
\usepackage{siunitx}
\usepackage{booktabs}
\usepackage{enumitem}
\usepackage{multirow}
\usepackage{subcaption}
\usepackage{booktabs}
\usepackage{tabularx}
\usepackage{multirow}
\usepackage{placeins}

\newcommand{\xihan}[1]{{\color{black}{#1}}}

\newcommand{\etal}{\textit{et al.}\xspace}

\begin{document}

\title{Decoding SEC Actions: Enforcement Trends through Analyzing Blockchain litigation using LLM-based Thematic Factor Mapping}

\settopmatter{authorsperrow=4}

\author{Junliang Luo}
\affiliation{%
  \institution{McGill University}
  \streetaddress{1 Th{\o}rv{\"a}ld Circle}
  \city{Montréal, Québec}
  \country{Canada}}

\author{Xihan Xiong}
\affiliation{%
  \institution{Imperial College London}
  \streetaddress{1 Th{\o}rv{\"a}ld Circle}
  \city{London}
  \country{United Kingdom}}

\author{William Knottenbelt}
\affiliation{%
  \institution{Imperial College London}
  \city{London}
  \country{United Kingdom}
}

\author{Xue Steve Liu}
\affiliation{%
 \institution{McGill University}
 \streetaddress{Rono-Hills}
 \city{Montréal, Québec}
 \country{Canada}}

\renewcommand{\shortauthors}{Anonymous Author(s)}

\begin{abstract}
The proliferation of blockchain entities (persons or enterprises) exposes them to potential regulatory actions (e.g., being litigated) by regulatory authorities.
Regulatory frameworks for crypto assets are actively being developed and refined, increasing the likelihood of such actions.
The lack of systematic analysis of the factors driving litigation against blockchain entities leaves companies in need of clarity to navigate compliance risks. This absence of insight also deprives investors of the information for informed decision-making.
This study focuses on U.S. litigation against blockchain entities, particularly by the U.S. Securities and Exchange Commission (SEC) given its influence on global crypto regulation.
Utilizing frontier pretrained language models and large language models, we systematically map all SEC complaints against blockchain companies from 2012 to 2024 to thematic factors conceptualized by our study to delineate the factors driving SEC actions.
We quantify the thematic factors and assess their influence on specific legal Acts cited within the complaints on an annual basis, allowing us to discern the regulatory emphasis, patterns and conduct trend analysis.

\end{abstract}
\vspace{-1cm}

%%
%% The code below is generated by the tool at http://dl.acm.org/ccs.cfm.
%% Please copy and paste the code instead of the example below.
%%



% \received{20 February 2007}
% \received[revised]{12 March 2009}
% \received[accepted]{5 June 2009}

\maketitle

\section{Introduction}

Current laws in jurisdictions such as the U.S., Canada and EU, are incorporating the concepts introduced by the blockchain economy into traditional financial regulations \cite{karisma2023blockchain, quintais2019blockchain, ducas2017security}.
The integration of concepts concerns the recognition and regulation of crypto service providers (CASPs), with a focus on compliance, consumer protection, and market integrity\cite{quintais2019blockchain} within established or new legal frameworks \cite{xiong2024global}. 
The enforcement of AML standards \cite{cox2014handbook} and applicable financial regulations is central to the regulation of cryptocurrency tokens across various jurisdictions.
%
% the operational security requirements by cybersecurity regulations \cite{didenko2020cybersecurity, allen2017sec}
These regulations are being developed to mitigate the risks associated with the pseudonymity and borderlessness of blockchain transactions, while leveraging their potential to reduce inefficiencies in securities settlement \cite{ducas2017security, collomb2019blockchain}.
Globally, regulation frameworks are being developed to address the regulation of digital assets, such as the \textit{Regulation on the Markets in Crypto-Assets (MiCA)} \cite{zetzsche2021markets} created by the European Commission, the SEC's \textit{Framework for “Investment Contract” Analysis of Digital Assets} \cite{framework2024} in the U.S.
These frameworks target establishing guidelines for the definition, classification and trading of digital assets (including crypto, the blockchain-based digital asset), maintaining financial stability and protecting investors.
Focusing on the present, these developing regulations have inevitably led to an increase in litigation in recent years as crypto assets challenge existing financial legal norms \cite{mcgurk2024risks}.
Crypto asset disputes often involve technical issues, asymmetric information, and impacts on investors and the broader socio-economic \cite{gabuthy2023blockchain, insights2023blockchain, notheisen2019blockchain}.
The disputes also interconnected with the market volatility triggered by blockchain activities and the ethical considerations of technology deployment that may influence public trust and participation in blockchain systems \cite{lianos2019regulating}.
Given the significant influence of U.S. financial regulations on global crypto assets regulation, this study focuses on U.S. litigation.
In the U.S., the SEC is actively pursuing lawsuit cases to ensure regulatory compliance and the protection of investors under the existing security laws.
The Commodity Futures Trading Commission (CFTC) also engages in litigation, regulating the crypto assets classified as commodities to ensure compliance with the \textit{Commodity Exchange Act}.
Given the distinct regulatory roles of the SEC and CFTC, this study focuses exclusively on SEC lawsuit cases, leaving CFTC-related litigation for potential future research.

\begin{figure*}[t!]
    \centering
    \includegraphics[width=0.998\textwidth]{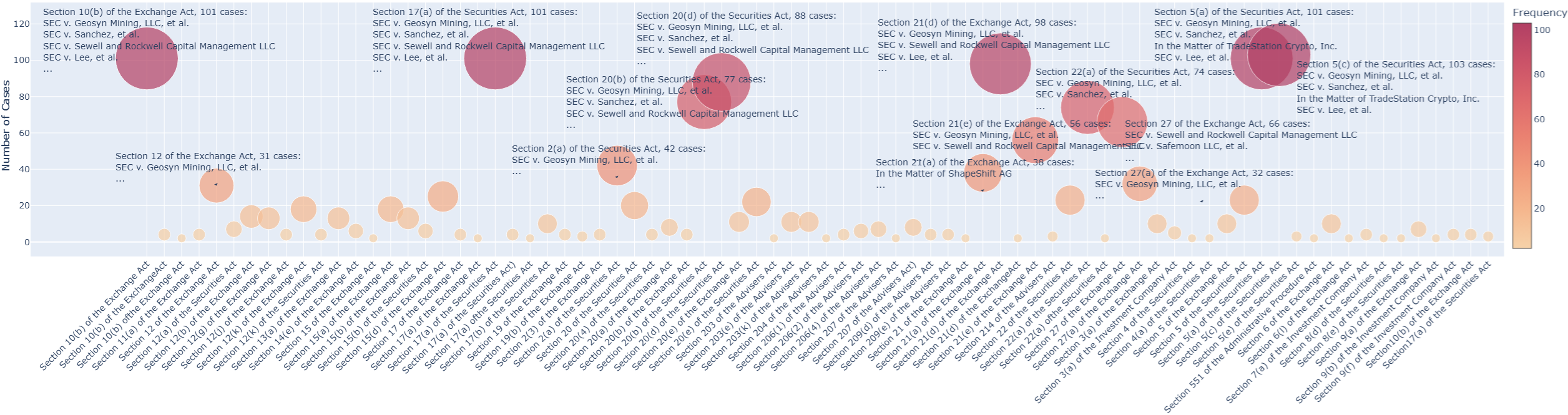}
    \caption{Association between the legal Acts and their corresponding lawsuit case groups. Larger circles in size indicate larger number of cases mentioning the same legal Act. Instances of lawsuit cases of some large groups are also given. }
    \label{fig:all_acts_cases}
\vspace{-0.28cm}
\end{figure*}

The SEC published \textit{Crypto Assets and Cyber Enforcement Actions} \cite{seclawsuits}, a comprehensive collection of SEC lawsuits against crypto entities from 2012 to 2024.
Given the collection, we obtained the complaints for all the cases. 
The complaints enumerate a constrained set of Acts (legal statutes enacted by legislative bodies that establish regulatory requirements and prohibitions) in a few sentences, yet the comprehensive content of complaints contains extensive information.
For instance, the high-profile case of SEC v. Ripple Labs, Inc. \cite{secpressreleaseripple2020} cites \textit{5(a) and 5(c) of the Securities Act} \cite{securities1933securities}, prohibiting the sale of unregistered securities, but the complaint itself spans over four hundred segments containing considerable details (detailed in Section \ref{ts_mapping}). 
The absence of analysis translating complaint details into clear, interpretable insights complicates understanding.
%
% This leaves enterprises lacking clarity to navigate compliance risks and deprives investors of crucial insights necessary for informed decisions, which may inhibit the responsible growth and development of the crypto industry. 
%
This leaves enterprises, even those with legal awareness, struggling to navigate compliance risks due to the complexity of legal documents, obscuring the full regulatory picture. Additionally, investors lack the insights needed to understand regulatory trends, potentially obstructing the growth of the crypto market.
To address this, we tackle explaining the underlying litigation drivers by employing quantitative analytics and modeling to systematically extract and interpret the latent factors triggering the litigation.
The research questions comprise three focused inquiries: How can we conceptualize a reasonable categorization for limited types of factors within the complaints, termed thematic factors, to delineate the factors driving SEC enforcement actions? 
How can these critical factors be extracted and quantified using existing machine learning language models, transforming legal text into measurable factors? 
How do these quantified thematic factors map to specific regulatory Acts, and what does this reveal about enforcement trends and regulatory priorities over time, enabling us to infer the form of activities most likely to precipitate SEC litigation.
We initiated the study by conceptualizing a set of thematic factors inspired by multiple sources (detailed in section \ref{thematic_semantic_factors}).
To extract and quantify the thematic factors in complaints, we proposed a method to assign each indexed segment (e.g. in Figure \ref{fig:overview_flow}) a corresponding thematic factor. 
The method leveraged recent advances in pretrained language models (PLMs) and large language models (LLMs).
The PLM transformed each segment of the SEC complaints into a vector, embedding their contextual representations within a semantic space.
Each of these vectors was mapped to the thematic factor by its similarity to the embedding vectors of LLM-generated seed sentences, each anchored to a specific thematic factor, and produced using the same PLM making the same semantic space.
Subsequently, we employed a Generalized Linear Model (GLM) to estimate the coefficients of these thematic factors with respect to specific legal Acts cited within the complaints on an annual basis. 
We conducted a detailed examination of these coefficients and identified the trends of regulatory emphases.
To the best of our knowledge,  we are the first to systematically analyze SEC complaints for blockchain entities from a data-driven perspective.
Our contribution provides CASPs and investors with a language model-based approach to systematically analyze SEC lawsuit complaints to extract and interpret regulatory trends, assisting their compliance strategies and decision-making. 
By uncovering and quantifying the driving factors behind SEC enforcement actions, we provide valuable insights into past enforcement patterns and potential future regulatory priorities.
This work serves as a resource for the blockchain community and the broader crypto industry, helping to clarify regulation and supporting the responsible development of the sector.
% added with the suggested contributions.
%
The conclusions extrapolated include:
\begin{itemize}[itemsep=2pt,topsep=-1pt,parsep=0pt,leftmargin=11pt]
    \item Practices detrimental to investors, such as fraud offering and misappropriation of consumer funds, remain a consistent and primary trigger for SEC enforcement actions across all years regardless of market conditions. It can be reasonably inferred that any entity engaging in such misconduct will continue to be subject to SEC litigation in the future.
    \item During market surges (e.g., crypto surges in 2017-18 \& 2021), the SEC's focus sharpened on the financial scale of companies' operations. Enterprises handling high market valuations of crypto asset, particularly those involved in unregistered securities offerings, are more likely to be scrutinized for non-compliance with registration mandates under the Securities Act. Entities offering assets during market surges involving large amounts of money face elevated risks of SEC litigation for unregistered offerings.
    \item The SEC's enforcement has expanded to cover a broader range of areas post 2020 such as tender offers, mandatory disclosures, and annual or quarterly reporting. Misuse of funds by key individuals was notably more prevalent before 2018, and celebrity promotions occurred in separate years, but it is reasonable to infer that the SEC continues to monitor such activities closely.
\end{itemize}

\section{Preliminary}
This section provides an overview of the crypto regulatory context, the SEC crypto assets litigation and the complaint data sources.

\vspace{-0.15cm}
\subsection{Crypto Regulatory Context}
The global regulation for cryptocurrencies varies across jurisdictions~\cite{xiong2024global,blandin2019global}, with regions implementing specialized frameworks such as the EU's Markets in Crypto-Assets (MiCA) regulation to oversee activities related to Crypto-Asset Service Providers (CASPs)
Other jurisdictions integrate cryptocurrency regulation within their existing legal frameworks. For example, in the U.S., the regulation of cryptocurrencies is incorporated into the existing securities and commodities regulatory frameworks. 
This divergence in regulatory approaches presents the challenges in achieving a unified global stance on cryptocurrency.
In the U.S., the SEC and the CFTC are the primary agencies responsible for overseeing cryptocurrencies. 
If a crypto asset qualifies as a security under the Howey Test~\cite{henning2018howey, hayes_crypto}, the asset falls under the jurisdiction of the SEC and must comply with the Securities Act of 1933 \cite{securities1933securities} and the Securities Exchange Act of 1934 \cite{securities1934securities}. 
Typically if a crypto asset does not fit the legal definition of a security under the Howey Test, the asset is classified as a commodity and regulated by the CFTC and must adhere to the Commodity Exchange Act of 1936 \cite{as1937commodity}. 

\subsection{SEC Crypto Assets Lawsuit Cases}
The SEC has actively pursued legal actions against various crypto entities in recent years due to violations of securities laws. 
One frequent violation is the failure to register with the SEC, which can manifest in several forms, such as unregistered brokers, traders, and clearing agencies operating without proper authorization.
The legal actions were against companies with cryptocurrency coin (token) offerings such as Ripple (XRP),  Telegram (TON), Binance (BNB), etc., alleging violations of federal securities laws regarding unregistered securities offerings \cite{beincrypto}.
Also, exchanges like Binance, Coinbase, Kraken, and Gemini have faced lawsuits for operating as unregistered brokers, traders, and clearing agencies offering unregistered securities \cite{cointelegraph2023}.
For instance, Binance and Coinbase have been accused of several violations, such as operating without proper registrations, mismanaging customer funds, and lacking adequate trading controls \cite{rcoincodecap}.
Fraudulent activities are another common violation that the SEC targets.
A notable case is the SEC's lawsuit against BitConnect, which alleged a two billion USD fraud involving a crypto investment scheme.
Market manipulation cases, such as the SEC's litigation against The Hydrogen Technology Corporation, et al., for manipulating the trading volume and price of the `Hydro' crypto asset.
Lawsuit cases are documented and accessible through Public Access to Court Electronic Records (PACER) \footnote{https://pacer.uscourts.gov/}, which provides access to case and docket information including all documents filed with the court, case-related transcripts, and orders from the court, culminating in a final judgment which resolves the legal issues in question and may include remedies such as penalties or directives to cease certain activities.
In our study, we focus on complaints of all these cases from SEC \textit{Crypto Assets and Cyber Enforcement Actions} \cite{seclawsuits} as the complaints are publicly available irrespective of the ongoing judicial status, and they provide a window into the obscured litigation elements and the SEC’s adaptive regulatory measures regarding blockchain assets. 

\section{Dataset}
\label{dataset}
The dataset \textit{Crypto Assets and Cyber Enforcement Actions} \cite{seclawsuits} consists of 226 distinct lawsuit cases filed against companies and or individuals from January, 2012, to July, 2024. 
The fields detail the action name, description, and date filed, and the link to the full complaint.
These cases involve accusations of unregistered securities offerings, account intrusions, insider trading, market manipulation, regulated entities, public company disclosure, and trading suspensions (official categories).
We obtained the complaints of all these cases.
In each complaint, the text is organized using segment indices. 
The indices number specific paragraphs within the complaint document to organize the content logically.
The logical organization typically presents an introduction, background details, factual allegations, legal charges, and the relief sought.
We processed the complaint of each case by segmenting each complaint into distinct segments by the indices.
Statistically, the derived SEC complaints dataset consists of a vocabulary size of 39,441 and an average word count of 6,042 per case. Each case is structured into averagely 79 segments, with an average segment length of 106 words.

\section{Thematic Semantic Mapping}
\label{ts_mapping}
\begin{figure}[t!]
    \centering
    \includegraphics[width=0.479\textwidth]{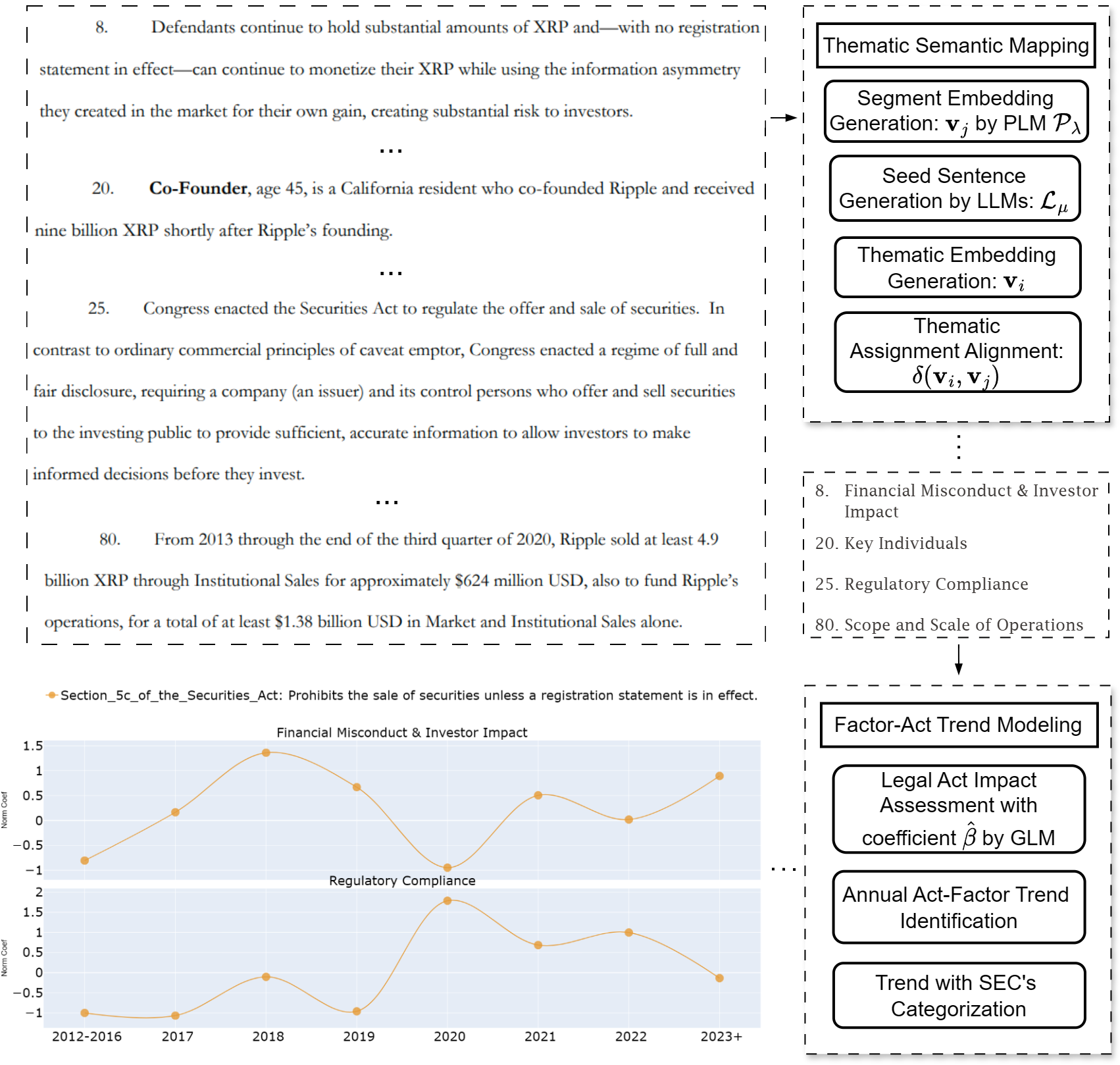}
    \caption{Workflow of thematic factor mapping and Act-factor trends modeling. Models are input with multiple complaints, SEC v. Ripple Labs shown as an illustrative example.}
    \label{fig:overview_flow}
\end{figure}

We extracted all legal Acts and associated group of lawcuit cases mention the same Acts presented in Figure \ref{fig:all_acts_cases}.
For instance, the complaint of the case: SEC v. Ripple Labs, Inc. mentioned \textit{Sections 5(a) and 5(c) of the Securities Act (of 1933)}, which prohibits the sale and delivery of unregistered securities (In this case, XRP) unless a registration statement is in effect; \textit{Section 20(b) of the Securities Act}, which holds individuals who control others liable for violations of the Securities Act.
The complaint mentions also the other Acts that allow the SEC to investigate violations of the securities laws, and provide the SEC with the power to enforce its provisions.
However, the SEC v. Ripple Labs exampled a complaint that encompasses extensive details, subdivided into over four hundred indexed segments.
Since the Acts provide only limited information on legal requisites, comprehending complaints through a method that can extract segment-level factors from complaints to gain a comprehensive understanding of the overall progression of blockchain assets is needed.
We demonstrate the whole workflow in Figure \ref{fig:overview_flow}.

\subsection{Thematic Factors}
\label{thematic_semantic_factors}
We initiate our workflow by operationalizing the litigation drivers using thematic factors as quantifiable proxies for potential regulatory triggers.
These thematic factors are drawn from publicly available complaints involving litigation by the SEC against various blockchain companies and entities, ranging from financial discrepancies to compliance breaches—and are hypothesized to directly correlate with litigative risks that precipitate SEC interventions. 
Drawing on insights from multiple publications and articles including Carmona et al. (2023) \cite{Carmona_Wolff_Levy_Yeyati_2023}, Goforth (2021) \cite{goforth2021regulation}, and Hennelly (2022) \cite{hennelly2022cryptic}, we hypothesized the thematic factors as the following:

\begin{itemize}[leftmargin=1pt,label={}]  
    \item \textbf{\textsc{Financial Misconduct \& Investor Impact}}: Focuses on the financial integrity and the direct impact on investors.
    \item \textbf{\textsc{Regulatory Compliance}}: Examines the adherence of companies to essential securities laws and regulatory mandates.
    \item \textbf{\textsc{Promotion \& Misrepresentation}}: Concentrates on the promotion and representation of information to investors.
    \item \textbf{\textsc{Scope \& Scale of Operations }}: Considers the extent and scale, specifically monetary values, of the company’s operations.
    \item \textbf{\textsc{Technological Risks}}: Represents the vulnerabilities within the technological deployments of the application.
    \item \textbf{\textsc{Key Individuals}}: Probes the roles and responsibilities of core company figures in legal compliance and possible misbehaviours.
\end{itemize}

\subsection{Mapping Method}
Some pretrained language models (PLM) can produce embedding vectors given the sentences that capture the contextual and semantic meanings within the text \cite{min2023recent}.
In our case, we apply the PLMs on each segment of all lawsuit cases of the dataset.
The segment-level embedding vectors present specific positions within the semantic space, situating the segment texts in a high-dimensional context representing their semantic meanings. 
The segment-level embedding vectors will be input to a model.
In such a model we employ a large language model (LLM) to map the embedding vector to a thematic semantic factor by synthesizing embedding vectors produced using LLM as a generative synthesizer to generate seed sentences targeting specific thematic factors.
The functions are defined as follows:
$\mathcal{L}_{\mu}$ is an LLM generative synthesizer to map $\mathbb{N}$ (factors in \ref{thematic_semantic_factors}) prompts to text outputs.
$\mathcal{P}_{\lambda}$ is an PLM to convert the input text into a $n$-dimension semantic embedding space.
$\mu$ and $\lambda$ indicate a particular kind of models.
\begin{equation*}
\mathcal{P}_{\lambda}: \text{Text} \rightarrow \mathbb{R}^n, \enspace \mathcal{L}_{\mu}: \text{Prompts} \times \mathbb{N} \rightarrow \text{Text}
\end{equation*}
$\mathcal{L}_{\mu}$ generates a sentence $\mathbf{seed\_s}_i$ of length limit  $\ell $ from prompt $\mathbf{p}_i$.
\begin{equation*}
\mathbf{seed\_s}_i = \mathcal{L}_{\mu}(\mathbf{p}_i, \ell), \quad \mathbf{v}_i = \mathcal{P}_{\lambda}(\mathbf{seed\_s}_i)
\end{equation*}
$\mathcal{P}_{\lambda}(\mathbf{t})$ converts text $\mathbf{t}$ into an n\text{-dimension embedding vector} $\mathbf{v}$.
\begin{equation*}
\mathbf{v}_j = \mathcal{P}_{\lambda}(\mathbf{s}_j), \quad \forall \mathbf{s}_j \in \text{Segments } \mathbf{S}
\end{equation*}
The seed sentences generated are assuredly aligned with designated thematic factors, and will be subsequently mapped into the identical embedding space produced by the same PLM as the segments.
Then each segment embedding $\mathbf{v}_j$ will be aligned with the thematic semantic factor $\mathcal{F}$ of its most similar seed sentence embedding $\mathbf{v}_i$ using the distance function $\delta$.
\begin{equation*}
\mathcal{F}(s_j) \leftarrow \mathcal{F}(\mathbf{seed\_s}_i) \quad \text{where} \quad i = \underset{i}{\operatorname{argmin}} \, \delta(\mathbf{v}_i, \mathbf{v}_j)
\end{equation*}
The aforementioned process is the mapping method performing the alignment of text segments to their thematic factors by leveraging the generated seed sentences and their embeddings.

\begin{table*}[h!]
    \caption{Thematic Factor IDs, names, and corresponding prompts}
    \label{tab:thematic_factors}
    \resizebox{0.978\textwidth}{!}{%
    \begin{tabular}{clp{16cm}}
        \toprule
        \textbf{Factor ID} & \textbf{Factor Name} & \textbf{Prompt} \\
        \midrule
        0 & Financial Misconduct \& Investor Impact (FM) & Focus on mentions of how the complaints describe the impact on investors, such as specific harms or financial losses. This includes improper accounting practices and bribery, highlighting the importance of compliance with internal controls and transparency in financial dealings. \\
        1 & Regulatory Compliance (RC) & Focus on mentions of companies failing to comply with regulations, such as not properly registering securities or failing to disclose critical information, and how these shortcomings subject them to lawsuits, emphasizing the importance of adherence to established securities laws. \\
        2 & Promotion \& Misrepresentation (PM) & Focus on mentions of any misrepresentations, particularly on instances of asymmetric information and over-promotion. Detail how information was misleading, exaggerated, or deceptively presented, offering insights into subtle forms of fraud. \\
        3 & Scope and Scale of Operations (SO) & Focus on mentions of how the scope and scale of the company’s operations are described. Pay attention to numeric facts such as the amount of money involved, the number of investors, and the geographic reach of operations. \\
        4 & Technological Risks (TR) & Focus on mentions of specific technological vulnerabilities or failures, such as inadequacies in the blockchain technology itself, security breaches, or technical misrepresentations. \\
        5 & Key Individuals (KI) & Focus on mentions of key individuals within the company. Observe how their actions, statements, and roles might reveal individual culpability or highlight leadership issues that contribute to legal violations. \\
        \bottomrule
    \end{tabular}
    }
\end{table*}

\begin{table*}[htbp!]
\caption{Comparison of normalized evaluation scores $R_i$, where $i$ corresponds to a factor ID. Results are presented with the score of each factor (under its abbreviation factor name as listed above) across various PLMs and LLMs.}
\label{tab:experiment_results}
\centering
\resizebox{0.999\textwidth}{!}{%
\renewcommand{\arraystretch}{1.08}
\begin{tabular}{@{}lccccccccccccccccccccccc@{}}
\toprule
\multicolumn{1}{c}{\multirow{2}{*}{\textbf{PLMs}}} & \multicolumn{7}{c}{GPT-4} & \multicolumn{1}{l}{} & \multicolumn{7}{c}{Meta Llama 3 (70B)} & \multicolumn{1}{l}{} & \multicolumn{7}{c}{Gemini} \\ \cmidrule(lr){2-8} \cmidrule(lr){10-16} \cmidrule(l){18-24} 
\multicolumn{1}{c}{} & FM & RC & PM & SO & TR & KI & M.V &  & FM & RC & PM & SO & TR & KI & M.V &  & FM & RC & PM & SO & TR & KI & M.V \\ \cmidrule(r){1-8} \cmidrule(lr){10-16} \cmidrule(l){18-24} 
all-MiniLM-L6-v2 & .414 & .650 & .451 & .372 & .580 & .459 & \textbf{.488 ± .106} &  & .396 & .446 & .525 & .331 & .523 & .454 & .446 ± .075 &  & .311 & .565 & .464 & .332 & .524 & .448 & .441 ± .102 \\
all-MiniLM-L12-v2 & .310 & .639 & .436 & .388 & .557 & .446 & .463 ± .118 &  & .337 & .463 & .452 & .399 & .505 & .491 & .441 ± .063 &  & .268 & .440 & .411 & .317 & .518 & .434 & .398 ± .091 \\
all-mpnet-base-v1 & .348 & .684 & .446 & .319 & .609 & .398 & .467 ± .147 &  & .348 & .533 & .478 & .380 & .529 & .471 & .457 ± .077 &  & .215 & .548 & .412 & .257 & .530 & .429 & .399 ± .137 \\
all-mpnet-base-v2 & .355 & .655 & .503 & .308 & .578 & .423 & .470 ± .133 &  & .324 & .507 & .436 & .380 & .571 & .504 & .454 ± .091 &  & .297 & .571 & .433 & .310 & .571 & .492 & .446 ± .122 \\
sentence-t5-base & .405 & .620 & .420 & .305 & .588 & .434 & .462 ± .119 &  & .347 & .437 & .405 & .348 & .601 & .501 & .440 ± .098 &  & .319 & .617 & .433 & .310 & .561 & .488 & .455 ± .125 \\
sentence-t5-large & .378 & .712 & .490 & .289 & .619 & .477 & \textbf{.494 ± .154} &  & .392 & .601 & .508 & .378 & .570 & .463 & \textbf{.485 ± .091} &  & .321 & .691 & .436 & .332 & .623 & .440 & \textbf{.474 ± .152} \\
gtr-t5-base & .383 & .694 & .437 & .502 & .649 & .425 & \textbf{.515 ± .128} &  & .361 & .504 & .478 & .436 & .577 & .387 & \textbf{.457 ± .080} &  & .338 & .679 & .376 & .353 & .561 & .457 & \textbf{.461 ± .135} \\
gtr-t5-large & .358 & .660 & .445 & .348 & .629 & .452 & .482 ± .133 &  & .399 & .585 & .516 & .352 & .545 & .493 & \textbf{.482 ± .089} &  & .318 & .676 & .404 & .315 & .593 & .510 & \textbf{.469 ± .149} \\ \bottomrule
\end{tabular}
}
\label{tab:metrics}
\end{table*}

\subsection{Experiment}
To assess the functioning of the proposed semantic embedding mapping, it is essential to develop a proper evaluation metric to assess the aligned thematic factors of segments, given that the task diverges from conventional downstream classification.
Subsequently, we conduct empirical evaluations applying a customized performance metric across a range of LLMs and PLMs.
This is to test LLMs' varying generalization capabilities in generating contextually appropriate thematic seed sentences, and the differing capacities of PLMs to capture semantic meaning, particularly within the specialized vocabulary and sentence characteristic of the complaints.

\subsubsection{PLM and LLM Models}
We selected the PLM models from the MiniLM, MPNet, Sentence-T5, and Generalizable T5 Retriever (GTR) families because they are compatible with Sentence Transformers \footnote{https://www.sbert.net/} and are known for their efficiency and performance in semantic search tasks \cite{wang2020minilm, song2020mpnet, ni2022sentence, ni2021large}.
We utilized the LLMs models ChatGPT-4 \cite{openaichatgpt}, Meta Llama 3 (70B) \cite{metallama}, and Gemini \cite{geminigoogle} due to their advances attributed to their extensive parameter numbers.
The prompts used to generate the seed sentences are detailed in Table \ref{tab:thematic_factors} beneath.

\subsubsection{Evaluation Metric}
Given the lack of the ground truth labels for each segment, we adopt an automated scoring metric using anchored lexical assessment without reliance on extensive labels.
The anchored lexical assessment is implemented via employing NER GLiNER model \cite{zaratiana2023gliner} to extract and score specialized terms that are anchored to each thematic semantic factor.
We compare the relevance of terms linked to the defined factor against terms from other factors within the segments.
Considering each thematic semantic factor $i$, the model extracts entities $E_{s_i}$ from all segments $S_i$ aligned with that factor. 
$$
sc_i^{(j)} = \frac{1}{\mathbb{N}} \sum_{s \in S_i} \sum_{e \in E_{s_i}} w_e \cdot \delta_{f_i}(e)
$$
The score for each factor $i$ is computed as above, where $\mathbb{N}$ is the total number of factors, $w_e$ is the confidence score output from the GLiNER model as the weight of entity $e$, and 
$\delta_{f_i}(e) =  1 \text{ if entity } e \text{ aligns with factor } \text{else } 0 $.
For each factor $i$, we obtain a list of scores $[sc_i^{(0)}, sc_i^{(1)}, ..., sc_i^{(\mathbb{N})}]$, where each score is the average weighted count of the relevant anchored terms by GLiNER model.
Given this list of scores, we define a normalized score $R_i$ for the factor $i$:
$$
R_i = \frac{\sum_{j \neq i} \Delta(sc_i^{(i)}, sc_i^{(j)})}{\sum_{k=1}^{\mathbb{N}} \sum_{j \neq k} \Delta(sc_i^{(k)}, sc_i^{(j)})}
$$
The difference between two two scores $sc_i^{(i)}$ and $sc_i^{(j)}$ is given by:
$$
\Delta(sc_i^{(i)}, sc_i^{(j)}) = \max(0, sc_i^{(i)} - sc_i^{(j)})
$$
The rationale is that for segments aligned with the factor $i$, a higher number of extracted anchored terms for factor $i$ compared to other factors indicates a good alignment, whereas high counts for both $i$ and other factors suggest inferior alignment.
Therefore we focus on calculating the positive difference as defined in the above difference function $\Delta$.
The normalized score calculates the normalized positive differential contributions of segments aligned with one factor, for anchored terms of that factor compared to others, ascending positively from no to maximal dominance in the 0-1 range.

\subsubsection{Results}
We conducted experiments using a segment length $\ell$ equivalent to the average length of segments (in section \ref{dataset}), and a set of one hundred sentences (testings of additional sentence volumes may be considered in future studies) using various GLMs and LLMs. 
The experiment results are presented in Table \ref{tab:experiment_results}.
The experimental results indicate that the ChatGPT-4 model obtained overall average scores compared to other LLMs. 
The MiniLM-based models report a stable metric distribution with minimal variance. 
In contrast, the gtr-t5 variants display the highest mean performance (as the top mean scores shown in bold), for example, with mean scores of 0.515 for the base model. 
The sentence-t5-large demonstrates a mean score of 0.494 but a higher variance among factors.
The evaluation result indicates that segments consistently show a relatively positive contribution in the non-marginal order toward their aligned thematic factors. 
This suggests that extracted terms associated with the segment's aligned factor are markedly more prevalent compared to terms from other factors, indicating a functioning thematic distinction for the alignments.

\section{Analysis}
This section analyzes the coefficient modeling between thematic factors and specific Acts, focusing on analyze the trend in the results and the alignment of these trends across various categories.

\subsection{Modeling Trends in Thematic Factor and Act Correlations}
The segment-level thematic factors were obtained and presented their efficacy.
To further uncover the research question of SEC priorities and regulatory trends over time, we conduct analysis to discern which thematic factors are most predictive of certain types of SEC legal Acts during a specific year.
We use the thematic factors produced by the PLM \textit{gtr-T5-base} and LLM \textit{GPT-4}, as they demonstrated comparatively better performance than all other models evaluated in our experiments.
We employed a Generalized Linear Model (GLM) with a logit link function to quantitatively assess the influence of thematic factors on the probability of SEC enforcement actions over successive years by the interpretable coefficients the GLM produced.
Denote our dataset $\{X_i, y_i, t_i\}_{i=1}^n$, where $X_i$ represents the thematic factors associated with the segments of the $i$-th complaint, $y_i$ denotes the presence of specific enforcement Acts, and $t_i$ indicates the year of the filing date.
$X_i = (x_{i1}, x_{i2}, \ldots, x_{ip})^\top$ is a vector of thematic factor proportions for the $i$-th complaint, $x_{ij}$ is the proportion of factor \(j\) in complaint $i$, $y_i \in \{0, 1\}$ is a binary indicator for the presence of a specific Act. 
The GLM for binary response with a logit link function is formulated as:
\begin{align*}
\text{logit}(P(y_i = 1 \mid X_i)) = \beta_0 + \sum_{j=1}^p \beta_j x_{ij} = \ln \left(\frac{P(y_i = 1 \mid X_i)}{1-P(y_i = 1 \mid X_i)}\right),
\end{align*}
% \vspace{-1cm}
%
where $\beta = (\beta_0, \beta_1, \ldots, \beta_p)^\top$ is the vector of coefficients for the GLM model, $\eta_i = \beta_0 + \sum_{j=1}^p \beta_j x_{ij}$ as the linear predictor, and $\mu_i = \frac{1}{1+e^{-\eta_i}}$ as the expected value of $y_i$ using the logistic function. 
The vector of coefficients $\beta$ is estimated to maximize the likelihood of observing the given data.
$\beta_0$ is the intercept term.
The model output denoted as $\hat{\beta}$, is the estimated value of the coefficient vector $\beta$, indicating the strength and direction of the influence of each factor on the probability of a legal Act.
The vectors of coefficients are Gaussian standardized, enabling categorizing the impact levels into three categories across different acts and time periods.
These coefficients demonstrate the alignment between thematic factors and specific regulatory Acts, confirming the relevance of these factors in existing legal Act in a certain year. 
The coefficients also offer an interpretable view of enforcement trends. 
The results are summarized in Table \ref{tab:trend_yearly}.
%
% These coefficients allow us to discern patterns in enforcement strategies, transform static legal references into dynamic indicators of regulatory emphasis.
%

\begin{table*}[htbp!]
\caption{Coefficients output from the GLM for the thematic factor and legal Act pairs. The table lists the maximum coefficient, the year of its occurrence, and categorizes the coefficient as high $\bullet$, moderate $\circ$, or low $\cdot$ on an annual basis. 2012-2016 are merged due to few cases; 2024 into 2023 for being incomplete. The top three Act-factor pairs based on the frequency of high coefficients are demonstrated. These metrics reflect trends in SEC enforcement actions over the analyzed period.}
\label{tab:experiment_results}
\centering
\resizebox{0.98\textwidth}{!}{%
\renewcommand{\arraystretch}{0.7}
\begin{tabular}{@{}llcccccccccc@{}}
\toprule
\multicolumn{1}{c}{\multirow{2}{*}{Act}} & \multicolumn{1}{c}{\multirow{2}{*}{ Thematic Factor}} & \multirow{2}{*}{Max Coef} & \multirow{2}{*}{Max Coef Yr} & \multicolumn{8}{c}{Annual Coef} \\
\multicolumn{1}{c}{} & \multicolumn{1}{c}{} &  &  & 2012-2016 & 2017 & 2018 & 2019 & 2020 & 2021 & 2022 & 2023+ \\ \midrule
Section 10(b) of the Exchange Act & Financial Misconduct \& Investor Impact & 1.899 & 2018 & \(\bullet\) & \(\bullet\) & \(\bullet\) & \(\circ\) & \(\cdot\) & \(\bullet\) & \(\bullet\) & \(\bullet\) \\
Section 17(a) of the Securities Act & Financial Misconduct \& Investor Impact & 1.970 & 2022 & \(\bullet\) & \(\bullet\) & \(\bullet\) & \(\bullet\) & \(\cdot\) & \(\bullet\) & \(\bullet\) & \(\circ\) \\
Section 5 of the Securities Act & Financial Misconduct \& Investor Impact & 1.408 & 2023 & \(\bullet\) & \(\cdot\) & \(\cdot\) & \(\bullet\) & \(\cdot\) & \(\circ\) & \(\bullet\) & \(\bullet\) \\
Section 14(e) of the Exchange Act & Regulatory Compliance & 1.550 & 2016 & \(\bullet\) & \(\bullet\) & \(\cdot\) &  &  & \(\cdot\) & \(\cdot\) &  \\
Section 13(a) of the Exchange Act & Regulatory Compliance & 1.156 & 2020 &  &  & \(\cdot\) & \(\cdot\) & \(\bullet\) & \(\bullet\) & \(\circ\) & \(\cdot\) \\
Section 12(g) of the Exchange Act & Regulatory Compliance & 1.156 & 2020 &  &  & \(\cdot\) & \(\cdot\) & \(\bullet\) & \(\cdot\) & \(\cdot\) &  \\
Section 17(b) of the Securities Act & Promotion \& Misrepresentation & 1.678 & 2023 &  &  & \(\cdot\) & \(\cdot\) & \(\cdot\) & \(\bullet\) & \(\bullet\) & \(\bullet\) \\
Section 206(4) of the Advisers Act & Promotion \& Misrepresentation & 1.067 & 2018 &  &  & \(\bullet\) &  &  & \(\cdot\) & \(\cdot\) & \(\bullet\) \\
Section 12(k) of the Exchange Act & Promotion \& Misrepresentation & 1.851 & 2018 &  & \(\cdot\) & \(\bullet\) & \(\bullet\) &  &  &  & \(\bullet\) \\
Section 5(a) of the Securities Act & Scope and Scale of Operations & 1.816 & 2016 & \(\bullet\) & \(\cdot\) & \(\bullet\) & \(\cdot\) & \(\cdot\) & \(\cdot\) & \(\bullet\) & \(\circ\) \\
Section 5(c) of the Securities Act & Scope and Scale of Operations & 1.816 & 2016 & \(\bullet\) & \(\cdot\) & \(\bullet\) & \(\cdot\) & \(\cdot\) & \(\cdot\) & \(\bullet\) & \(\circ\) \\
Section 15(a) of the Exchange Act & Scope and Scale of Operations & 1.582 & 2021 & \(\cdot\) &  & \(\bullet\) & \(\cdot\) &  & \(\bullet\) & \(\circ\) & \(\circ\) \\
Section 13(a) of the Exchange Act & Technological Risks & 1.330 & 2023 &  &  & \(\cdot\) & \(\cdot\) & \(\cdot\) & \(\bullet\) & \(\cdot\) & \(\bullet\) \\
Section 15(b) of the Exchange Act & Technological Risks & 1.408 & 2019 & \(\circ\) &  & \(\cdot\) & \(\bullet\) & \(\cdot\) & \(\circ\) & \(\bullet\) & \(\circ\) \\
Section 3(a) of the Exchange Act & Technological Risks & 1.876 & 2017 & \(\circ\) & \(\bullet\) & \(\bullet\) &  &  & \(\cdot\) &  & \(\circ\) \\
Section 10(b) of the Exchange Act & Key Individuals & 1.904 & 2020 & \(\bullet\) & \(\circ\) & \(\cdot\) & \(\bullet\) & \(\bullet\) & \(\cdot\) & \(\circ\) & \(\cdot\) \\
Section 20(a) of the Exchange Act & Key Individuals & 1.178 & 2022 &  &  & \(\bullet\) &  &  &  & \(\bullet\) & \(\cdot\) \\
Section 12(a) of the Securities Act & Key Individuals & 1.270 & 2019 &  &  & \(\bullet\) & \(\bullet\) & \(\cdot\) &  & \(\cdot\) &  \\ \bottomrule
\end{tabular}
}
\label{tab:trend_yearly}
\end{table*}

Considering the overall results, multiple thematic factors exhibit high coefficients with sections such as \textit{Section 20 of the Securities Act, Section 20(e)}, \textit{22(a) of the Securities Act}, and \textit{Section 21(d), 21(a) of the Exchange Act}, which presents the SEC's broad enforcement powers to investigate violations, seek injunctions, impose penalties, and enforce compliance with securities laws. 
Since these Acts are primarily related to the general enforcement mechanisms rather than the specific thematic factors and legal provisions, these sections are not central to our analytical focus.
Focusing on the distinct correlation, \textit{Financial Misconduct \& Investor Impact} shows consistently high coefficients with \textit{Section 5 and Section 17(a) of the Securities Act}, indicating strong enforcement against unregistered securities offerings and fraud. 
\textit{Regulatory Compliance} shows strong associations with \textit{Section 12(k), 12(g) of the Exchange Act} suggesting heightened regulatory scrutiny on registration requirements. 
\textit{Promotion \& Misrepresentation} exhibits rising coefficients linked to \textit{Section 17(b) of the Securities Act}, \textit{Section 206(4) of the Advisers Act}, reflecting intensified enforcement against misrepresentation and undisclosed promotions.
\textit{Scope and Scale of Operations} shows coefficients with \textit{Section 12a, 12g of the Exchange Act}, indicating operational requirements, such as requiring companies to register securities traded on national exchanges.
\textit{Technological Risks} present high coefficients related to \textit{Section 3(a) of the Exchange Act}, reflecting the attention to how crypto asset trading terms are integrated with the key concepts and definitions from the Exchange Act.
\textit{Key Individual}s is notably linked to \textit{Section 10(b) of the Exchange Act} for \textit{Section 206 of the Advisers Act} for prohibiting fraud by investment advisers, representing a focus on individual misconduct and high-profile executive cases.
Beyond specific correlations, our research objective emphasizes analyzing the temporal Act-factor trend to uncover shifts in the SEC's enforcement priorities.
To conduct the analysis, each Act-factor pair was ranked based on the number of years where coefficients surpassed 1.0, categorizing them as high. Values over 0.5 were deemed moderate and all others low. 
We disregarded negative coefficients since they complicate the interpretation of direct impacts or influences, focusing instead on positive coefficients to provide the insights of direct regulatory emphasis.
From 2012 to 2016, the SEC focused heavily on the market, investors, and regulation mentioning \textit{Section 10(b) of the Exchange Act} to combat securities fraud and deceit. The emphasis on \textit{Section 14(e) of the Exchange Act} regulating tender offers, which allows shareholders in a private company to sell some or all of their shares. 
At this very early period, we saw enforcement under \textit{Section 5 of the Securities Act}, which addresses the registration of securities, stressing the agency's commitment to ensuring that all securities offerings were duly registered.
In 2017, the SEC began to shift its focus toward technological risks and further enhancing regulatory compliance, as evidenced by increased enforcement of \textit{Section 3(a) of the Exchange Act}, related to broker-dealer registration. This focus was likely a response to the rise of digital trading platforms and of the cryptocurrency price in 2017.
Both \textit{Section 10(b) of the Exchange Act} and \textit{Section 17(a) of the Securities Act} remain high coefficients with the \textit{Financial Misconduct \& Investor Impact factor}, reflecting the SEC continued addressing securities fraud and deceptive practices.
Following the rise of trading platforms, the SEC presented the first time focus on promotion in 2018, concerning fraudulent promotional activities and misleading financial endorsements, such as celebrities who promoted ICOs on social media without disclosing the fact.
\textit{Promotion \& Misrepresentation} factor highly correlated to \textit{Section 17(b) of the Securities Act}, which prohibits the promotion of securities without full disclosure of compensation received for such promotion, accompanied by \textit{Section 206(4) of the Advisers Act} prohibits false statements by investment advisers, reflecting the SEC's strategy to curb misleading promotions and endorsements on social media platforms. 
The period likely witnessed an SEC's response to the burgeoning impact of social media and celebrity endorsements, where undisclosed financial incentives could lead to conflicts of interest. 
Meanwhile, the coefficient linking misrepresentation to \textit{Section 12(k) of the Exchange Act} is associated with addressing short sale regulations in the market.
\textit{Sections 5(a) and 5(c) of the Securities Act} exhibited high coefficients for \textit{Scope and Scale of Operations}, reflecting the SEC's intensified focus on compliance with securities registration during the period of market expansion.

\begin{table*}[ht]
    \centering
    \caption{SEC cases related to crypto and cyber assets (by year and SEC's categorization)}
    \resizebox{0.97\textwidth}{!}{%
    \renewcommand{\arraystretch}{0.98}
    \begin{tabular}{@{}llcl@{}}
    \toprule
    \textbf{Year} & \textbf{Category} & \textbf{Cases} & \textbf{Notable Cases} \\ \midrule
    2016 & Account Intrusions & 3 & SEC v. Mustapha: Hacked brokerage accounts for unauthorized trades\\
    2017 & Account Intrusions & 1 & SEC v. Willner: Hacked brokerage accounts to manipulate stock prices \\
    2022 & Account Intrusions & 1 & SEC v. Mohamed, et al.: Fraudulent scheme hacking brokerage accounts \\
    2016- & Crypto Assets & 6 & SEC v. Garza, et al.: Bitcoin mining Ponzi scheme; SEC v. Shavers: Bitcoin Ponzi scheme with high returns \\
    2017 & Crypto Assets & 5 & SEC v. PlexCorps, et al.: Fraudulent ICO scheme; SEC v. REcoin Group Foundation, LLC, et al.: Fraud with ICOs backed by real estate \\
    2018 & Crypto Assets & 16 & SEC v. Longfin Corp., et al.: Fraudulent trading activities \\
    2019 & Crypto Assets & 17 & SEC v. Kik Interactive Inc.: \$100 million unregistered securities offering; SEC v. ICOBox, et al.: Illegal \$14 million securities offering \\
    2020 & Crypto Assets & 23 & SEC v. Meta 1 Coin Trust, et al.: Conducting a fraudulent initial coin offering of unregistered digital asset securities \\
    2021 & Crypto Assets & 18 & SEC v. BitConnect, et al.: \$2 billion fraud with crypto lending platform; SEC v. LBRY, Inc.: Unregistered offering of digital asset securities \\
    2022 & Crypto Assets & 23 & SEC v. Wahi, et al.: Insider trading charges against a former Coinbase manager \\
    2023 & Crypto Assets & 36 & SEC v. Hex et al.: \$1 billion raised fraudulently; SEC v. Coinbase, Inc.: Operating as an unregistered securities exchange \\
    2024 & Crypto Assets & 6 & SEC v. Geosyn Mining, LLC: Fraudulent securities offering; SEC v. Sanchez, et al.: \$300 million Ponzi scheme \\
    2016- & Hacking/Insider Trading & 4 & SEC v. Dubovoy, et al.: Newswire hack for insider trading; SEC v. Hong, et al.: Law firm hack for insider trading \\
    2019 & Hacking/Insider Trading & 1 & SEC v. Ieremenko, et al.: Hacked SEC's EDGAR system for illegal trading \\
    2021 & Hacking/Insider Trading & 2 & SEC v. Kliushin, et al.: Profited from stolen earnings announcements \\
    2022 & Hacking/Insider Trading & 1 & SEC v. Dishinger, et al.: Insider trading with Equifax breach data \\
    2023 & Hacking/Insider Trading & 1 & O. Kuprina: Alleged hacking scheme targeting the SEC's Electronic Data Gathering, Analysis, and Retrieval (EDGAR) system \\
    2017 & Market Manipulation/False Tweets/Fake Websites/Dark Web & 1 & SEC v. Murray: Manipulated stocks with false SEC filings \\
    2018 & Market Manipulation/False Tweets/Fake Websites/Dark Web & 1 & SEC v. Burns: Manipulating stock price with false EDGAR filing \\
    2020 & Market Manipulation/False Tweets/Fake Websites/Dark Web & 2 & SEC v. Sotnikov, et al.: Lured investors with fake websites \\
    2021 & Market Manipulation/False Tweets/Fake Websites/Dark Web & 5 & SEC v. Trovias: Sold "insider tips" on the Dark Web; SEC v. Gallagher: Stock manipulation using false Twitter posts \\
    2022 & Market Manipulation/False Tweets/Fake Websites/Dark Web & 6 & SEC v. EmpiresX, et al.: \$40 million fraud with fake daily profits; SEC v. Parrino: Market manipulation with false rumors \\
    2023 & Market Manipulation/False Tweets/Fake Websites/Dark Web & 1 & SEC v. Patel: False rumors for illicit profits \\
    2016- & Market Manipulation/False Tweets/Fake Websites/Dark Web & 3 & SEC v. PTG Capital Partners LTD, et al.: False SEC filings to manipulate prices \\
    2018 & Public Company Disclosure and Controls & 2 & Altaba Inc., f.d.b.a Yahoo! Inc.: Concealment of a massive data breach \\
    2021 & Public Company Disclosure and Controls & 2 & First American Financial Corporation: Poor cybersecurity disclosures \\
    2022 & Public Company Disclosure and Controls & 1 & NVIDIA Corporation: Inadequate disclosures on cryptomining impact \\
    2023 & Public Company Disclosure and Controls & 2 & Solarwinds Corp.: Fraud and internal control failures \\
    2016- & Regulated Entities – Cybersecurity Controls and Safeguarding Customer Information & 2 & Morgan Stanley Smith Barney LLC: Failed to safeguard customer data \\
    2018 & Regulated Entities – Cybersecurity Controls and Safeguarding Customer Information & 1 & Voya Financial Advisors: Cybersecurity failures and data protection issues \\
    2019 & Regulated Entities – Cybersecurity Controls and Safeguarding Customer Information & 2 & Virtu Americas LLC: Violations in dark pool operation compliance \\
    2021 & Regulated Entities – Cybersecurity Controls and Safeguarding Customer Information & 4 & KMS Financial Services, Inc.: Cybersecurity failures exposing client data \\
    2022 & Regulated Entities – Cybersecurity Controls and Safeguarding Customer Information & 4 & Morgan Stanley Smith Barney LLC: Failure to protect customer PII; J.P. Morgan Securities LLC: Identity theft prevention deficiencies \\
    2023 & Regulated Entities – Cybersecurity Controls and Safeguarding Customer Information & 1 & Options Clearing Corporation: Non-compliance with SEC-approved stress testing rules \\
    2024 & Regulated Entities – Cybersecurity Controls and Safeguarding Customer Information & 1 & Intercontinental Exchange, Inc.: Failure to timely inform SEC of cyber intrusion \\
    2016- & Trading Suspensions & 1 & In re Imogo Mobile Technologies Corp.: Trading suspended due to questions about claims of a secure mobile Bitcoin platform \\
    2017 & Trading Suspensions & 7 & The Crypto Co.: Trading was suspended due to concerns about the accuracy of information regarding insider plans to sell shares \\
    2018 & Trading Suspensions & 9 & PDX Partners, Inc.: Trading suspended due to concerns over information accuracy and business operations. \\
    2019 & Trading Suspensions & 1 & Bitcoin Generation, Inc.: Trading suspended due to concerns about the impact of stock promotional activities \\
    2021 & Trading Suspensions & 2 & Long Blockchain Corp.: Trading suspension due to failure in timely filings \\
    2022 & Trading Suspensions & 1 & American CryptoFed DAO LLC: Halted registration of digital tokens \\ \bottomrule
    \end{tabular}
    }
\label{tab:all_case_categories}
\end{table*}

\begin{table*}[htbp!]
\centering
\caption{SEC Categories, averaged alignment score for each category, percent of cases with high score (Cse. pct.). The top three most contributive pairs are demonstrated. Category names are abbreviated for clarity.}
\Large
\resizebox{1.0\textwidth}{!}{
\begin{tabular}{@{}lcclll@{}}
\toprule
\textbf{Category} & \textbf{Avg. Score} & \textbf{Cse. pct.} & \multicolumn{2}{c}{\textbf{Most Contributive Pairs}} & \textbf{Act Description} \\ \midrule
Crypto Assets & 1.477 & .662 & \begin{tabular}[c]{@{}l@{}}Section 5(a) of the Securities Act\\ Section 5(c) of the Securities Act\end{tabular} & \begin{tabular}[c]{@{}l@{}}Scope and Scale of Operations\\ Scope and Scale of Operations\end{tabular} & \begin{tabular}[c]{@{}l@{}}Prohibits the sale or offer of securities in interstate commerce without an effective registration statement filed. \\ Prohibits the sale of securities unless a registration statement is in effect.\end{tabular} \\
Account Intrusions & 1.647 & .750 & \begin{tabular}[c]{@{}l@{}}Section 17(a) of the Securities Act\\ Section 10(b) of the Exchange Act\end{tabular} & \begin{tabular}[c]{@{}l@{}}Financial Misconduct \& Investor Impact\\ Financial Misconduct \& Investor Impact\end{tabular} & \begin{tabular}[c]{@{}l@{}}Prohibits fraud and deceit in the offer or sale of securities.\\ Prohibits manipulative and deceptive practices in connection with the purchase or sale of securities.\end{tabular} \\
Hacking/Insider Trading & 1.371 & .444 & \begin{tabular}[c]{@{}l@{}}Section 21(d) of the Exchange Act\\ Section 20(b) of the Securities Act\end{tabular} & \begin{tabular}[c]{@{}l@{}}Key Individuals\\ Key Individuals\end{tabular} & \begin{tabular}[c]{@{}l@{}}Grants the SEC authority to seek court orders to enforce compliance with the Exchange Act.\\ Addresses the jurisdiction and venue for legal actions under the Securities Act.\end{tabular} \\
Market Manipulation & 1.406 & .631 & \begin{tabular}[c]{@{}l@{}}Section 21(d) of the Exchange Act\\ Section 17(a) of the Securities Act\end{tabular} & \begin{tabular}[c]{@{}l@{}}Key Individuals\\ Financial Misconduct \& Investor Impact\end{tabular} & \begin{tabular}[c]{@{}l@{}}Grants the SEC authority to seek court orders to enforce compliance with the Exchange Act.\\ Prohibits fraud and deceit in the offer or sale of securities.\end{tabular} \\
Regulated Entities & 0.766 & .263 & \begin{tabular}[c]{@{}l@{}}Section 203(e) of the Advisers Act\\ Section 203(k) of the Advisers Act\end{tabular} & \begin{tabular}[c]{@{}l@{}}Regulatory Compliance\\ Regulatory Compliance\end{tabular} & \begin{tabular}[c]{@{}l@{}}Allows the SEC to censure, place limitations on, or suspend or revoke the registration of investment advisers.\\ Provides the SEC with authority to impose sanctions on investment advisers for violations.\end{tabular} \\
\begin{tabular}[c]{@{}l@{}}Public Company \\ Disclosure and Controls\end{tabular} & 1.059 & .574 & \begin{tabular}[c]{@{}l@{}}Section 12 of the Exchange Act\\ Section 13(a) of the Exchange Act\end{tabular} & \begin{tabular}[c]{@{}l@{}}Technological Risks\\ Technological Risks\end{tabular} & \begin{tabular}[c]{@{}l@{}}Requires securities to be registered with the SEC to be traded on national exchanges.\\ Requires issuers to file annual and quarterly reports with the SEC.\end{tabular} \\
Trading Suspensions & 0.673 & .173 & \begin{tabular}[c]{@{}l@{}}Section 12(k) of the Exchange Act\\ Section 8(d) of the Securities Act\end{tabular} & \begin{tabular}[c]{@{}l@{}}Regulatory Compliance\\ Regulatory Compliance\end{tabular} & \begin{tabular}[c]{@{}l@{}}Addresses short sale regulations and the reporting of certain trading activities.\\ Addresses the process and conditions under which the SEC can issue a stop order against a registration statement\end{tabular} \\ \bottomrule
\end{tabular}
}
\label{tab:a_scores}
\end{table*}

From 2018 to 2020, the SEC continued to address security registration while also targeting key individuals under \textit{Section 10(b) of the Exchange Act and Section 20(a)} concerning controlling persons. 
This demonstrated the SEC's commitment to holding executives and other high-profile figures accountable for misconduct, reinforcing the principle of individual accountability within corporate governance.
In 2020, the SEC focused on regulatory compliance under \textit{Section 13(a) of the Exchange Act}, concerning periodic financial reporting, for companies' transparent disclosure practices. 
In 2021, the SEC’s enforcement efforts emphasized \textit{Scope and Scale of Operations}, particularly correlated with \textit{Section 15(a) of the Exchange Act}, which governs broker-dealer activities. 
The SEC’s strategic intent is to oversee the operations of large-scale firms and ensure their compliance with the registration and operational provisions.
This year was characterized by a bull market, with rising crypto prices, which may have influenced the SEC's scrutiny on broker-dealer activities and high-volume market operations.
From 2022 to 2023, the SEC intensified its focus back on financial misconduct and investors, notably mentioning \textit{Section 10(b), 17(a) of the Securities Act} to address fraud in securities offerings. 
Simultaneously in 2023, the SEC focused on enforcing \textit{Section 17(b) of the Securities Act} and \textit{Section 206(4) of the Advisers Act}, showing high coefficients for promotion and misrepresentation to target fraudulent promotions and misleading financial advisement, while \textit{Section 12(k) of the Exchange Act} was also mentioned to authorize trading suspensions and protect market integrity.

\subsection{Trend Alignment across SEC's Categories}
We calculate an alignment score for each case to quantify how closely the factors and Acts in a case align with the overall (all-inclusive) trend of Act-factor coefficients obtained in the previous section.
The alignment scores of individual cases are calculated to examine their distribution across the SEC's official categorization. 
This scores help understand whether the SEC's categorization aligns with or diverges from broader enforcement patterns, i.e., whether certain categories, exhibit consistent enforcement patterns, or present more variability suggesting diverse regulatory challenges.
The alignment score $S_c$ of a category $c$ is defined as follows, where each case (compliant) $c$ is associated with a set of Acts $A_c$ and factor proportions $X_c$, as well as a specific year $y_c$. 
$$
S_c = \frac{\sum_{i \in A_c} \sum_{j \in X_c} p_j \cdot C^{y_c}_{ij}}{\sum_{i \in A_c} \sum_{j \in X_c} \left(\frac{C^{y_c}_{ij}}{n}\right)}
$$
$p_j$ is the proportion of factor $x_i$ in the case, and $C^{y_c}_{ij}$ is the overall correlation coefficient between Act $i$ and factor $j$ for the corresponding year $y_c$. 
The denominator assumes an equal distribution of factors across Acts normalizing the score. 
A high alignment score ($S_c \geq 1.0$) suggests that the specific combination of factors and Acts in the case corresponds more closely to the overall trends than a hypothetical scenario where factors are evenly distributed.
The average alignment scores across categories are presented in Table \ref{tab:a_scores} with the percentage of cases with high scores for the cases of each category.
The table identifies the most contributive Act-factor pairs for each category. i.e., a pair of Act and factor for a given year that has the highest correlation coefficients of a category.
For instance, the \textit{Crypto Assets} category with the most cases prominently features \textit{Section 5(a) and Section 5(c) of the Securities Act} paired with \textit{Scope and Scale of Operations} indicating a persistent regulatory focus on unregistered securities offerings and the amount of money involved for crypto assets for all the lawsuit cases in this category. 
The higher average scores and larger percentage of high-scoring cases in categories like \textit{Crypto Assets} and \textit{Market Manipulation} suggest strong conformity to the overall trends. Conversely, lower scores in categories like \textit{Regulated Entities} and \textit{Trading Suspensions} may indicate more diverse or evolving challenges, suggesting at diverse complexities.
The relation between cases in each category and the overall trend indicates that \textit{Crypto Assets} and \textit{Market Manipulation} consistently align with global trends patterns due to well-defined enforcement practice, whereas others, such as \textit{Regulated Entities} and \textit{Trading Suspension} show more variability in litigation reasoning, suggesting diverse regulatory challenges.

\section{Related work}
This section reviews literature relevant to crypto regulation, crypto litigation and the techniques for analyzing legal documents.

\xihan{
\begin{description} [leftmargin=*]
    \item[Crypto Regulation.] 
    The regulation of cryptocurrencies has been a subject of global interest. For instance, Blandin~\etal~\cite{blandin2019global} analyzed regulatory frameworks for cryptoassets in 23 jurisdictions. The study found significant variability in regulatory approaches. 
    %and highlights challenges such as inconsistent terminology and regulatory gaps. 
    %
    Transitioning from the global perspective, many scholars have focused on the legal framework specific to the United States. Hughes~\cite{hughes2017cryptocurrency} examined various federal and state-level enforcement actions, the challenges of regulating decentralized digital currencies, and the legal definitions provided by various U.S. agencies. 
    The paper concluded that while significant efforts have been made to regulate the cryptocurrency market, the existing patchwork of laws often creates confusion and inconsistency.  
    Moffett~\cite{moffett2022cftc} discussed the regulatory challenges and jurisdictional conflicts between the CFTC and the SEC concerning cryptocurrencies. The article suggested a dual regulatory framework and emphasized the need for cooperation between the CFTC and SEC.

    \item[Crypto Litigation.] Research has also investigated the wider scope of crypto litigation beyond SEC actions. 
    Ghodoosi~\cite{ghodoosi2022crypto} provided an empirical analysis of crypto-related cases litigated in the United States using the dataset from the Morrison Cohen Crypto Litigation Tracker \footnote{https://www.morrisoncohen.com/insights/the-morrison-cohen-cryptocurrency-litigation-tracker}. 
    This study examined the number of cases, types of disputes, and causes of actions involving cryptocurrencies, tokens, exchanges, and decentralized autonomous organizations (DAO). 
    It revealed that while early cases were dominated by securities litigation from the ICO boom, there has been a significant shift towards private law claims involving contracts and torts. 
    This trend, termed the ``private law pivot,'' suggests future crypto litigation will focus on more complex private law issues. 
    Yahya and Pecharsky~\etal~\cite{yahya2022crypto} provided an overview of crypto litigation from 2020, categorizing cases by causes of action and highlighting the prevalence of fraud, breach of contract, and regulatory infractions. 
    Their findings underscored the diversity of legal challenges faced by the crypto industry, from fraudulent investment schemes to criminal prosecutions.
    
    \item[Natural Language Processing in Legal Analysis.] The application of Natural Language Processing (NLP) techniques to analyze legal documents has proven valuable in understanding regulatory and litigation trends \cite{hassan2021addressing}. 
    % For instance, Chalkidis~\etal~\cite{chalkidis2020legal} explored adapting BERT models for legal tasks. 
    For instance, Chalkidis~\etal~\cite{chalkidis2020legal} explored adapting BERT models for legal tasks such as multi-label text classification of EU laws, binary and multi-label classification of European Court of Human Rights cases, and named entity recognition in US contracts.
    The study compared using BERT out of the box, further pre-training on legal texts, and pre-training from scratch with a legal-specific vocabulary and found that domain-specific pre-training improved performance.
    The authors introduced LEGAL-BERT, a family of BERT models optimized for legal texts, showing improved results in the aforementioned tasks. 
    Merchant~\etal~\cite{merchant2018nlp} used Latent Semantic Analysis (LSA) for legal text summarization. 
    The method involves pre-processing the text, creating a term-document matrix, and applying Singular Value Decomposition (SVD) to identify key sentences. 
    The method was tested using a dataset of legal judgments from Indian courts and achieved a decent average ROGUE-1 score.
    Jallan~\etal~\cite{jallan2019application} used the LexisNexis database \cite{lexisnexis_academic} to automatically extract and analyze cases from the past decade. 
    By applying the Latent Dirichlet Allocation (LDA) model, they identified and classified common themes and patterns in these cases. 
    The study concludes that the automated method can effectively identify broad patterns in construction-defect litigation. 
\end{description}
}

% Crypto regulation

% SEC Enforcement Actions

% Crypto litigation

% Natural Language Processing in Legal Analysis

% nlp study:

\section{Conclusion}
We approach a systematic investigation of the drivers behind SEC enforcement actions against blockchain and cryptocurrency entities.
The drivers are substantively delineated through thematic factors defined by our study conceptualized in response to previous publications.
By leveraging pretrained language models and large language models, we provide a data-driven semantic mapping method for quantifying these factors and assessing their impact on regulatory actions. 
By aligning thematic factors with regulatory Acts, we offer insights into the SEC's evolving priorities, providing the understanding of the SEC's adaptive focus annually. 

\bibliographystyle{ACM-Reference-Format}
\bibliography{sample-base}

% \section{Appendix}
% \begin{minipage}{\textwidth}
%     \centering
%     \includegraphics[width=1.00\textwidth]{figure/factor_act_year_new.png}
%     \captionof{figure}{Normalized coefficients of regulatory Acts across thematic factors from 2012 to 2023, output by the GLM models evolving the regulatory acts mentioned in more than two SEC enforcement actions against blockchain companies. Each colored marker represents a specific Act with legend presenting a brief description at a certain year indicating the coefficient.}
%     \label{fig:factor_act_year_all}
% \end{minipage}

\end{document}